\documentclass[journal]{IEEEtran}

\usepackage{hyperref}
\usepackage{amsmath}
\usepackage{amssymb}
\usepackage{caption}
\usepackage{bm}
\usepackage[ruled,norelsize]{algorithm2e}
\usepackage{float}
\usepackage{multirow}
\usepackage{flushend}
\ifCLASSINFOpdf
   \usepackage[pdftex]{graphicx}
\else
\fi
\newcommand{\norm}[1]{\left\lVert#1\right\rVert}

\begin{document}
%
\title{Unsupervised MKL in Multi-layer Kernel Machines}
%
%
%
\author{\IEEEauthorblockN{
        Akhil Meethal\IEEEauthorrefmark{1},
        Asharaf S\IEEEauthorrefmark{2},        
        Sumitra S\IEEEauthorrefmark{1}
        }\\
      \IEEEauthorblockA{\IEEEauthorrefmark{1} Department of Mathematics, Indian Institute of Space Science and Technology, Trivandrum}\\
      \IEEEauthorblockA{\IEEEauthorrefmark{2} School of CS \& IT, Indian Institute of Information Technology and Management - Kerala}
  }

\markboth{Journal of \LaTeX\ Class Files,~Vol.~14, No.~8, August~2020}%
{Shell \MakeLowercase{\textit{et al.}}: Bare Demo of IEEEtran.cls for IEEE Journals}
%



\maketitle

\begin{abstract}
Kernel based Deep Learning using multi-layer kernel machines(MKMs) was proposed by Y.Cho and L.K. Saul in \cite{saul}. In MKMs they used only one kernel(arc-cosine kernel) at a layer for the kernel PCA based feature extraction. We propose to use multiple kernels in each layer by taking a convex combination of many kernels following unsupervised learning strategy. Empirical study is conducted on \textit{mnist-back-rand}, \textit{mnist-back-image} and \textit{mnist-rot-back-image} datasets generated by adding random noise in the image background of MNIST dataset. Experimental results indicates that using MKL in MKMs earns better representation of the raw data and improves the classifier performance.
\end{abstract}

\begin{IEEEkeywords}
Deep Learning, Multiple Kernel Learning, Deep Kernel Learning
\end{IEEEkeywords}

%
\IEEEpeerreviewmaketitle

\section{Introduction}
\label{introduction}
%
%
%
%
\IEEEPARstart{}{R}epresentation Learning is one of the key area in machine learning, where intense research works are being done spanning over decades. The choice of the representation has a considerable influence in the performance of learning algorithms\cite{bengioRL}. Representation Learning is also considered as an effective mechanism for transferring the prior knowledge of humans to the machine learning system. The architecture of the Representation Learning system is broadly classified into deep and shallow architectures\cite{bengioAI}. Shallow architectures often produces strong task specific priors, whereas deep architectures can support broad priors which can favour a wide range of functions.

Deep Learning is a rapidly developing area among Representation Learning algorithms, which makes use of depth as a key criteria for producing good representations(hence they belong to deep architectures). In general, Deep Learning is a set of algorithms in machine learning that attempts to model high-level abstractions in data by using model architectures composed of multiple non-linear transformations\cite{bengioLDA}\cite{bengioRL}. Deep Learning methods are preferred over shallow ones in many complex learning tasks(such as computer vision, speech recognition etc.) due to: the wide range of functions that can be parameterized by composing weakly nonlinear transformations, the broad range of prior knowledge they can transfer to the learning system, the invariance being modelled by such systems against local changes in the input and their ability to learn more abstract concepts in an hierarchical fashion\cite{saul}\cite{bengioAI}.

Though the idea of using such deep architectures for learning representations is known to machine learning researchers, two main factors deterred the wide use of Deep Learning algorithms: the gradient-based methods often gets trapped in local minima which results in poor generalization capacity, training the networks with more than 2 or 3 layers was a big computational challenge. The \textit{greedy layerwise unsupervised pre-training} proposed by Hinton et.al in \cite{hintonDBN} is a breakthrough in machine learning research tackling the difficulty in training such deep models. This method proposes to learn a hierarchy of features one level at a time, following an unsupervised strategy\cite{bengioRL}. This greedy layerwise unsupervised pre-training is used to initialize the weights of the network(unsupervised pre-training stage) and then the weights are fine-tuned with the given label information(supervised fine-tuning stage).

The recent rise in computing power alleviated the second problem upto a considerable level. Efficient training methods are devised for learning networks with millions of parameters both on multi-core CPUs and GPUs\cite{gpu1}\cite{gpu2}. Scalable deep learning models are also developed in distributed cloud based enviornments\cite{graphlab}.

Armed with these tools Deep Learning witnessed a sudden boost since 2006, producing state-of-the-art results in several complex machine learning tasks like speech recognition\cite{speechBest}, visual object recognition\cite{imagenet}\cite{cifarBest}\cite{mnistBest}, natural language processing\cite{nlpBest} etc. However, most of the Deep Learning algorithms are developed on top of neural network based models\cite{lecunnCNN}\cite{hintonDBN} which shows a declining trend in the popularity of kernel machines in advanced learning tasks. Many researchers have been extensively studied the limitations kernel machines in such problems. The analysis shown in \cite{bengioAI} claims that the number of training examples required for a kernel machine with Gaussian kernel may grow linearly with the number of variations of the target function.

In \cite{saul} Cho et.al explored the concept of \textit{kernel methods based Deep Learning} by proposing Multi-layer Kernel Machines(MKMs). The architecture of MKMs consists multiple layers with each layer performing unsupervised feature extraction using kernel PCA\cite{kpca} followed by supervised feature selection by ranking features based on their mutual information with class labels. We tried to extend this architecture by taking combination of multiple kernels at each layer. In particular, instead of using a single kernel for the kernel PCA based feature extraction we used a convex combination of multiple kernels(Multiple Kernel Learning following unsupervised paradigm) at each layer.

The organization of this paper is as follows. In section \ref{mkms} we reviews the existing MKMs proposed by Cho et.al along with the kernel function introduced by them(arc-cosine kernels which can mimic the computations in large multilayer threshold networks). Section \ref{mlmkms} describes the proposed Multi-layer Multiple Kernel Learning(ML-MKL) framework for representation learning. Section \ref{experiments} describes the experimental results and some analysis done on this ML-MKL framework. Section \ref{conclusion} concludes this paper.


\section{Multi-layer Kernel Machines}
\label{mkms}
\begin{figure}[h]
  \centering
  \captionsetup{justification=centering,margin=0.1cm}
  \includegraphics[scale=0.3]{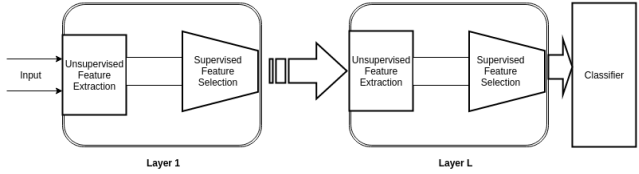}
  \caption{An MKM with L layers of transformations. Each layer consists of unsupervised feature extraction(using kernel PCA) followed by supervised feature selection.}
  \label{fig_mkm}
\end{figure}

In MKMs the non-linear transformations are induced by kernel functions stacked in a layerwise fashion. The architecture of MKMs are similar to that of neural network based deep learning machines, with unsupervised feature extraction(using Kernel PCA\cite{kpca}) followed by supervised feature selection in each layer. The figure \ref{fig_mkm} shows the architecture of an MKM consisting of L layers of non-linear transformations.

The supervised feature selection allows us to retain only those set of features that are most informative for the pattern recognition task. The number of features to be passed to the next layer(width of that layer) are found out using cross validation techniques. This procedure determines the architecture of the network in a greedy, layer-by-layer fashion. In their implementation\cite{saul} Cho et. al used exhaustive search in a particular range to determine the optimal set of features.
The output from the final layer can be passed to any classifier. Though any kernel can be used for the kernel PCA based feature extraction, in \cite{saul} they emphasized on using arc-cosine kernels due to: their similarity with computation in large neural networks, the multiple layers of kernel composition does meaningful computation only in case of arc-cosine kernels.

\subsection{Arc-cosine Kernels}
Let x, y be two inputs in $\mathbb{R}^d$. Define $\theta$ as the angle between them.
\[ \theta = cos^{-1}\left ( \frac{x\cdot y}{\left \| x \right \| \left \| y \right \|} \right ) \]
Then the kernel function computed by the arc-cosine kernel is
\begin{equation}
k_n(x,y) = \frac{1}{\pi}\left \| x \right \|^n \left \| y \right \|^n J_n(\theta) 
\end{equation}
where n is called the \textit{degree of the kernel} and
\[ J_n(\theta) = (-1)^n(sin\theta)^{2n+1} \left ( \frac{1}{sin\theta} \frac{\partial}{\partial \theta} \right )^n \left ( \frac{\pi-\theta}{sin\theta} \right ) \]

$J_n(\theta)$ for n=0, 1, 2 is computed as shown below.
\[ J_0(\theta) = \pi-\theta \]
\[ J_1(\theta) = sin\theta + (\pi-\theta)cos\theta \]
\[ J_2(\theta) = 3sin\theta cos\theta + (\pi-\theta)(1+2cos^2\theta) \]
for n=0, it takes the simple form
\[k_0(x,y) =  1- \frac{1}{\pi}cos^{-1}\left ( \frac{x\cdot y}{\left \| x \right \| \left \| y \right \|} \right )  \]
hence the name arc-cosine kernel is given. Since arc-cosine kernels do not have any continuous tuning parameters, cross-validation for finding the optimal degree value is not a tiresome job.

A kernel function is implicitly inducing a non-linear mapping from input $x$ to feature vector
$\phi(x)$(except the linear kernel). In effect, the kernel function computes the inner product of two input datapoints $x \textrm{ and } y$ in this induced feature space:
\[ k(x,y) = \phi(x) \cdot \phi(y) \]
Iteratively applying the mapping $\phi(\cdot)$ on the inputs $x \textrm{ and } y$ and then taking their inner product, we can obtain an L-layer kernel function as
\[ k^{(L)}(x,y) = \underbrace{\phi(\phi(\ldots \phi(x)))}_{\textrm{L times}} \textrm{ } \cdot \textrm{ } \underbrace{\phi(\phi(\ldots \phi(y)))}_{\textrm{L times}} \]
The intuition behind this in case of arc-cosine kernels is, if the base kernel $k(x,y) = \phi(x) \cdot \phi(y)$ can mimic the computation of a single-layer network, then the iterated mapping in $k^{(L)}(x,y)$ can mimic the computation of multi-layer network\cite{saul}. The L layer kernel function for arc-cosine kernels can be recursively computed as
\[ k^{(L+1)}_n(x,y) = \frac{1}{\Pi} \Bigg[ k^{(L)}(x,x) \textrm{ } k^{(L)}(y,y)\Bigg]^{\frac{n}{2}} J_n(\theta_n^{(L)}) \]
where $\theta_n^{(L)}$ is the angle between images of $x \textrm{ and } y$ in the feature space after L layer composition
\[ \theta_n^{(L)} = \textrm{cos}^{-1}\Bigg( k^{(L)}(x,y) \Bigg[ k^{(L)}(x,x) \textrm{ } k^{(L)}(y,y)\Bigg]^{\frac{-1}{2}} \Bigg) \]
for simplicity we have assumed that the arc-cosine kernels have the same degree $n$ at every layers of recursion. We can also use kernels of different degree at different layers.

\section{Multi-layer Multiple Kernel Machines}
\label{mlmkms}

Multiple Kernel Learning(MKL)\cite{mkl} aims at learning a convex combination of a set of predefined base kernles inorder to get a good kernel. The primary aim of MKL algorithm is to reduce the burden of choosing the optimal kernel(and hence automate the kernel parameter tuning) for the learning task. Also in practical scenarios instead of using a single kernel, a combination of multiple kernels give better results for pattern recognition, regression etc.

The concept of Multi-layer Multiple Kernel Learning has been explored by many authors\cite{2l_mkl}\cite{deep_mkl} in supervised learning settings. In particular the authors in \cite{2l_mkl} and \cite{deep_mkl} embedded the ML-MKL framework inside the SVM cost function, where instead of a single flat kernel a multi-layer kernel with many kernels in each layer is used. The authors used a combination of Convex Quadratic Programming(for finding coefficients of decision function, $\alpha$) and gradient descent(for getting the kernel weights $\mu$) in an alternating fashion to solve the resulting optimization task. Though the results are encouraging, the framework is not tested in large datasets studied extensively in Deep Learning literatures. Another limitation of the above approach is, the feature learning process is specific to a classifier.

In the KPCA based feature extraction stages, we were using only one kernel for the task. With the intuition that by using multiple kernels at each layer we will get more similarity information, we computed a convex combination of multiple kernels following the work in \cite{zhuang}. Since we are using kernels for unsupervised feature extraction, traditional MKLs following supervised paradigm cannot be used here.

The goal of an unsupervised multiple kernel learning task is to find an optimal linear combination of the $m$ kernel functions as, i.e, $k^*(\cdot \textrm{ , } \cdot) \in K_{conv}$, where $K_{conv}$ is defined as follows:
\[ K_{conv} = \Bigg\{ k( \cdot \textrm{ , } \cdot ) = \sum_{t=1}^m \mu_t k_t(\cdot \textrm{ , } \cdot) \textrm{ : } \sum_{t=1}^m \mu_t = 1, \mu_t \geq 0 \Bigg\} \]  
where $k_t$'s are the base kernels. Inorder to determine the optimality of a linear combination of kernels, we used the following quality criteria:

\begin{itemize}
\item A  good kernel should enable each training instances to be well reconstructed from the localized bases weighted by the kernel values. Formulating this requirment mathematically, for each $x_i$ we expect the optimal kernel should minimize the approximation error $\norm{x_i-\sum_j k_{ij}x_j}^2$.
\item A good kernel should induce kernel values that are coincided with the local geometry of the training data. This is equivalent to finding the optimal kernel that minimizes the distortion over all trainig data, computed as $\sum_{i,j}k_{ij} \norm{x_i-x_j}^2 $, where $k_{ij} = k(x_i, x_j)$.
\end{itemize}

In addition to this, the locality preserving principle is also exploited here by using a set of local bases for each $x_i \in X$ denoted as $B_i$. Instead of using the entire training data, we used these local bases to reconstruct the sample $x_i$ and compute the distortion as well. By fixing the size of the local bases to some constant $N_B$, we formulate the optimization problem of unsupervised MKL as follows.
\[ \min_{k \in K_{conv}} \frac{1}{2}\sum_{i=1}^n \norm{x_i - \sum_{x_j \in B_i} k_{ij}x_j}^2 + \gamma* \sum_{i=1}^n \sum_{x_j \in B_i} k_{ij} \norm{x_i-x_j}^2 \]

where $\gamma$ is a tuning parameter, which controls the tradeoff between the coding error and the locality distortion. Converting to matrix notations we will get the following optimization problem
\begin{equation}
  \min_{\mu \in \Delta,D} \frac{1}{2} \norm{X(I-K \circ D)}_F^2 + \gamma* \textrm{ tr }K\circ D\circ M(11^T)
  \label{obj_mat} 
\end{equation}
\[ \textrm{subject to } D \in \{0,1\}^{n \times n} \]
\[  \norm{d_i}_1 = N_B, i=1,2,\ldots,n \]
\[ \Delta = \Big\{\mu : \mu^T\textbf{1} = 1, \mu \geq 0 \Big\} \textrm{ and } \]
\[ [K]_{i,j} = \sum_{t=1}^m \mu_t k^t(x_i, x_j), 1\leq i,j \leq n \]

The matrix $D \in \{0,1\}^{n \times n}$ contains information about local bases of each $x_i$ as a column vector. In particular, each column vector $d_i \in \{0, 1\}^n $ in $D$ has a 1 at those points $j$ where, $x_j \in B_i$ and zero elsewhere(or $B_i = \{ x_j : d_j \neq 0 \} $). The matrix M is defined as
\[ [M]_{ij} = x_i^Tx_i + x_j^Tx_j - 2x_i^Tx_j \]

In equation \ref{obj_mat} the notation `$\circ$' denotes elementwise multiplication of two matrices, $\norm{\cdot}_F^2$ denotes the Frobenius norm of a matrix and `tr' denotes the trace of a matrix.

In their implementation(\cite{zhuang}) Zhuang et.al solved the optimization problem in two stages in an alternating fashion, first by solving for $\mu$ with a fixed $D$(using convex optimization) and then solving for $D$ by fixing $\mu$(using a greedy mixed integer programming formulation). Since we are using a many layer architecture, the alternating optimization strategy is too costly for us; so we choose to do the optimization across $\mu$ only by choosing $D$ beforehand. Specifically, the matrix $D$ is computed beforehand by taking $k$ nearest neighbours of $x_i$ from the training set and putting a one in those positions for $d_i$. Rest of the positions are filled with zeros. The resulting optimization problem will be  
\begin{equation}
  \begin{aligned}
  \min_{\mu \in \Delta} \frac{1}{2} \norm{X(I-K \circ D)}_F^2 + \gamma* \textrm{ tr }K\circ D\circ M(11^T)
  \end{aligned}
\end{equation}
\[ \textrm{subject to } D \in \{0,1\}^{n \times n}, \norm{d_i}_1 = N_B, i=1,2,\ldots,n \]
\[ \Delta = \Big\{\mu : \mu^T\textbf{1} = 1, \mu \geq 0 \Big\} \]

The objective function can be formulated as a Convex Qudratic Programming problem w.r.t to kernel weights $\mu$ as shown below(derivation of the objective function $J(\mu)$ is shown in Appendix \ref{derivation1}).
\begin{equation}
J(\mu) = \mu^T \Bigg( \sum_{t=1}^m \sum_{i=1}^n k_{t,i}k_{t,i}^T \circ d_i d_i^T \circ P \Bigg)^T \mu + z^T \mu 
\end{equation}
where $[z]_t = \sum_{i=1}^n (2 \gamma v_i \circ d_i - 2 p_i \circ d_i)^T \mathit{k}_{t,i} $, $P = X^TX$, and $\mathit{k}_{t,i} = \Big[ k^t(x_i, x_1), \ldots, k^t(x_i, x_n) \Big]^T $ is the $i^{th}$ column of the $t^{th}$ kernel matrix. $p$ and $v$ are columns of $P$ and $M$ corresponding to $x_i$ respectively.

\subsection{ML-MKL Algorithm}
\begin{figure*}
  \centering
  \captionsetup{justification=centering,margin=0.1cm}
  \includegraphics[width=0.8\textwidth,height=4.5cm]{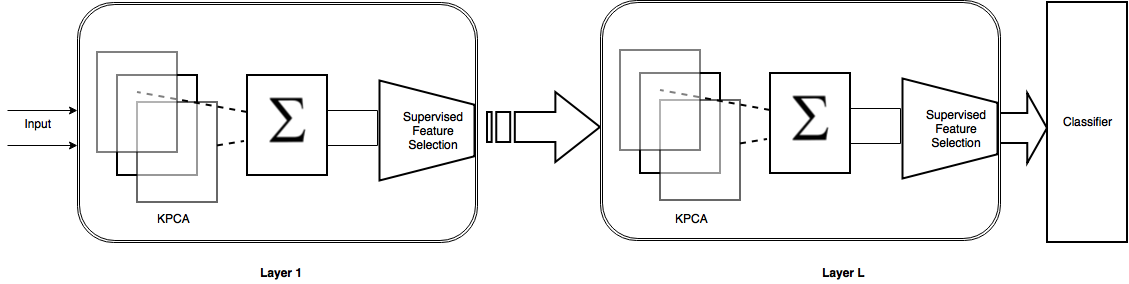}
  \caption{An ML-MKL with L layers of transformations. Each layer consists of many kernels for feature  extraction using kernel PCA and a supervised feature selection module.}
  \label{fig_mlmkl}
\end{figure*}

The architectue of the proposed ML-MKL framework is shown in figure \ref{fig_mlmkl}. It consists of many layers and in each layer the kernel PCA based feature extraction is performed using the combination of a set of predefined kernels. The dimensionality of the features thus obtained are reduced by using supervised feature selection techniques. The final output can be given to any classifier. Algorithm \ref{algo1} summarizes the proposed ML-MKL algorithm.

\begin{algorithm}
\caption{ML-MKL Algorithm}
\textbf{Input}: data X, true labels y, no. of layers L, base kernels for each layer $K_{base}^{(l)} = \{k_1^{(l)}, k_2^{(l)}, k_m^{(l)}\}$, $N_B$, $\gamma$\;
\textbf{Output}: kernel weights $\mu^l$ for each layer, predicted labels\;
1.Initialize $[M]_{ij} = x_i^Tx_i + x_j^Tx_j - 2x_i^Tx_j$, $\bm{D} = d_1, d_2, \ldots, d_n$ as row vectors, where $d_i = \{1 \textrm{ if } x_j \in B_i \textrm{ else } 0 \forall x_j \in X \}$, $\mu = \frac{1}{m}$, $\bm{P} = X^TX$\;
2.\For{each layer l}{
    a. $\bm{W} = \sum_{t=1}^m \sum_{i=1}^n k_{t,i}^{(l)}k_{t,i}^{(l)^{T}} \circ d_i d_i^T \circ P$\\
    b. $\bm{[z]}_t^l = \sum_{i=1}^n (2 \gamma v_i \circ d_i - 2 p_i \circ d_i)^T \mathit{k}_{t,i}^{(l)}$\\
    c. $\bm{\mu}^{*^{l}} = \mu^{l^{T}}W\mu^l + z^{l^{T}}\mu^l$\\
    d. $\bm{K}_{new} = \sum_{t=1}^m \mu_t^l * K_t^{(l)}$\\
    e. extract principal components with $\bm{K}_{new}$\\
    f. select most informative features for layer $l$($X_{new}$)\\
    g. $\bm{P} = X_{new} ^T X_{new}$
}
3.Give the final set of features to any classifier\;
\label{algo1}
\end{algorithm}

\section{Experiments}
\label{experiments}
\begin{figure}[h]
  \centering
  \captionsetup{justification=centering,margin=0.1cm}
  \includegraphics[scale=0.3]{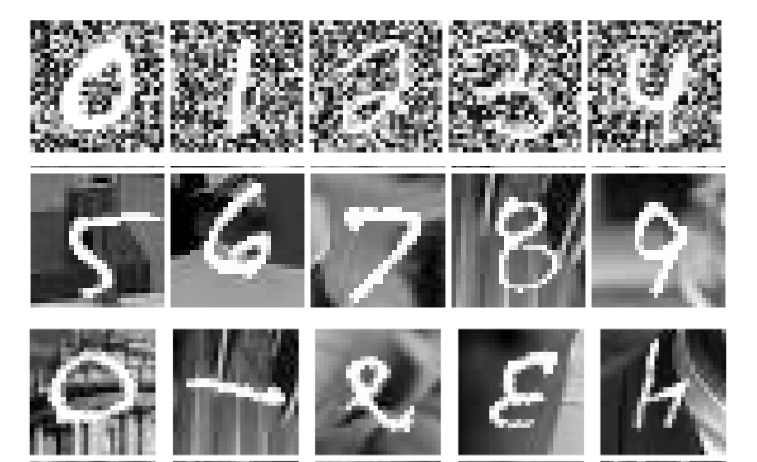}
  \caption{sample images from \textit{mnist-back-rand}(first row), \textit{mnist-back-image}(second row) and \textit{mnist-rot-back-image}(third row) datasets.}
  \label{samples}
\end{figure}

Empirical study is conducted on three datasets, created from MNIST dataset\cite{mnist} of handwritten digits by adding noise in the background(and by performing some transformations also). The \textit{mnist-back-rand} dataset was created by filling the image background with random pixel values, \textit{mnist-back-image} dataset by filling the image background with random image patches, and \textit{mnist-rot-back-image} is a rotated variant of \textit{mnist-back-image} where the rotation angle is generated uniformly between $0$ and $2\pi$. Each image is of size 28$\times$28 and each dataset contains 12000 training images and 50000 testing images. Figure \ref{samples} contains some samples from these the datasets.

In the training phase, for each layer we set apart 10000 datapoints for training the model and 2000 datapoints for cross validating kernel parameters and the tuning parameter $\gamma$. For the kernel PCA we choose 3000 datapoints from the training set randomly and repeated the test 5 times to get a mean value of the performance. The feature selection is done by performing a univariate statistical test and then ranking the features based on the test score, followed by selection of most informative features sorted according to the rank. For performing the univariate statistical test feature selection module available in the \textit{scikit-learn}\cite{scikit} library is used.

In the final classification stage SVMs with arc-cosine kernels are used. For a fair comparison with MKMs the same feature selection techique(univariate feature selection) and classifier(SVM with arc-cosine kernel) is used in our implementation of MKMs. Table \ref{tab_results} summarizes the results of our empirical study. 

Table \ref{tab_results} also contains best results obtained from other models\cite{hintonDBN} like SVM with RBF kernel(SVM RBF), SVM with polynomial kenel(SVM Poly), single hidden layer feed-forward neural network(NNet), DBN with 1 hidden layer(DBN-1), DBN with 3 hidden layer(DBN-3) and 3 hidden layer Stacked Autoassociator Network(SAA-3). The first three models comes under shallow architectures and the remaining are deep architectures. From the table, it can be observed that ML-MKL outperforms all the remaining models except Deep Belif Networks(DBN). Compared to DBN the architecture, parameter tuning and optimization etc. are fairly simple in ML-MKL models. The performance gap is very much considerable with shallow architectures(SVM RBF, SVM Poly, NNet) in case of \textit{mnist-back-rand} dataset.

\renewcommand{\arraystretch}{2.3}
\begin{table*}
\centering
\begin{tabular}{|c|c|c|c|c|c|c|c|c|}
  \hline
  \multirow{2}{*}{\textbf{Dataset}} & \multicolumn{8}{ |c| }{\textbf{Loss in Percentage}} \\
  \cline{2-9}
  & SVM RBF & SVM Poly & NNet & DBN-3 & SAA-3 & DBN-1 & MKM & \textbf{ML-MKL}\\
  \hline  
  \textit{mnist-back-rand} & 14.58 & 16.62 & 20.04 & \textbf{6.73} & 11.28 & 9.80 & 10.55 & 8.43$\pm 0.088$\\
  \hline
  \textit{mnist-back-image} & 22.61 & 24.01 & 27.41 & 16.31 & 23.00 & \textbf{16.15} & 21.39 & 20.92$\pm 0.092$\\
  \hline
  \textit{mnist-rot-back-image} & 55.18 & 56.41 & 62.16 & \textbf{47.39} & 51.93 & 52.21 & 51.61 & 51.21$\pm 0.811$\\
  \hline
\end{tabular}
\caption{Experimental Results of ML-MKL}
\label{tab_results}
\end{table*}
\renewcommand{\arraystretch}{1}

For the \textit{mnist-back-rand} dataset the best result is obtained with a model consists of 4 layers and in each layer 7 kernels are used. In particular, each layer consists of a mixture of one arc-cosine kernel and 6 gaussian kernels. For the \textit{mnist-back-image} dataset the best ML-MKL model obtained has 2 layers and 5 kernels in each layer. In each layer a mixture of one arc-cosine kernel and 4 polynomial kernels are used. In the case of \textit{mnist-rot-back-image} dataset the best result is fetched by a model having only one layer with 4 arc-cosine kernels in it. For all datasets the layer width is set to 150 features in each layer after cross validation. 

\begin{figure}
  \centering
  \captionsetup{justification=centering,margin=0.1cm}
  \includegraphics[width=0.52\textwidth,height=8.5cm]{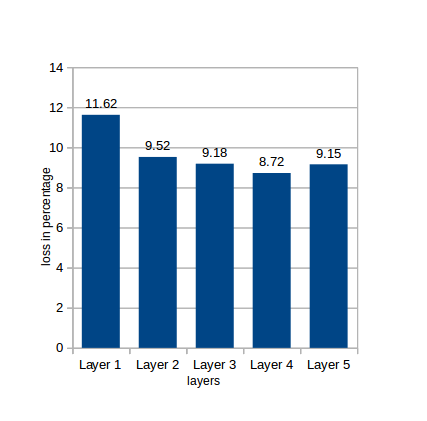}
  \caption{Change in classifier performance on \textit{mnist-back-rand} dataset when adding layers iteratively.}
  \label{fig_mbr_layers}
\end{figure}

\begin{figure}
  \centering
  \captionsetup{justification=centering,margin=0.1cm}
  \includegraphics[width=0.42\textwidth,height=8.5cm]{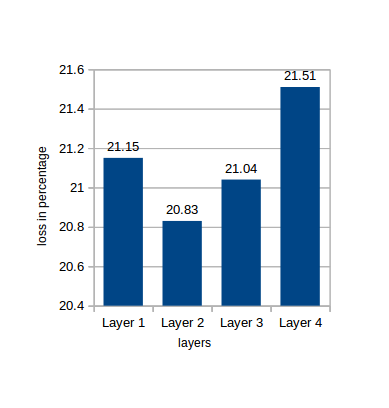}
  \caption{Change in classifier performance on \textit{mnist-back-image} dataset when adding layers iteratively.}
  \label{fig_mbi_layers}
\end{figure}

Figures \ref{fig_mbr_layers} and \ref{fig_mbi_layers} illustrates the variation in classifier performance on \textit{mnist-back-rand} and \textit{mnist-back-image} datasets respectively when layers are added iteratively to the ML-MKL model. The value shown for each layer is the best error rate obtained after tuning the kernel parameters. The parameters are choosen greedily for each layer(with the expectation that subsequent layers will learn more valuable features from the current one) and no fine-tuning is performed with respect to the entire architecture.

Tables \ref{back_rand_kw} and \ref{back_image_kw} shows the kernel weights of each kernel in the mixture at every layer for \textit{mnist-back-rand} and \textit{mnist-back-image} datasets respectively. In both cases $k_1$ is an acr-cosine kernel, and the remaining are Gaussian kernels for \textit{mnist-back-rand} dataset and Polynomial kernel for textit{mnist-back-image} dataset. The results in the table indicates that, in a multi-layer architecture the contribution of individual kernels is highly varying in each layer(no single kernel has complete dominance over all layers in the feature learning process). However the exact reason why some kernels are weighted more in a particular layer than the other kernels is not very clear.

\renewcommand{\arraystretch}{2.3}
\begin{table*}
\centering
\begin{tabular}{|c|c|c|c|c|c|c|c|}
  \hline
  \multirow{2}{*}{\textbf{layers}} & \multicolumn{7}{ |c| }{\textbf{Kernel Weights}} \\
  \cline{2-8}
  & $k_1$ & $k_2$ & $k_3$ & $k_4$ & $k_5$ & $k_6$ & $k_7$ \\
  \hline  
  \textit{Layer 1} & 0.2007 & 0.1331 & 0.1331 & 0.1332 & 0.1332 & 0.1333 & 0.1333\\
  \hline
  \textit{Layer 2} & 0.2711 & 0.1160 & 0.1181 & 0.1203 & 0.1225 & 0.1248 & 0.1271\\
  \hline
  \textit{Layer 3} & 0.1764 & 0.0999 & 0.1125 & 0.1266 & 0.1426 & 0.1607 & 0.1811\\
  \hline
  \textit{Layer 4} & 0.0598 & 0.0524 & 0.0747 & 0.1071 & 0.1547 & 0.2245 & 0.3269\\
  \hline
  \textit{Layer 5} & 0.0514 & 0.0493 & 0.0726 & 0.1069 & 0.1569 & 0.2295 & 0.3334\\
  \hline    
\end{tabular}
\caption{Kernel weights in each layer for the \textit{mnist-back-rand} dataset}
\label{back_rand_kw}
\end{table*}
\renewcommand{\arraystretch}{1}

\renewcommand{\arraystretch}{2.3}
\begin{table}
\centering
\begin{tabular}{|c|c|c|c|c|c|}
  \hline
  \multirow{2}{*}{\textbf{layers}} & \multicolumn{5}{ |c| }{\textbf{Kernel Weights}} \\
  \cline{2-6}
  & $k_1$ & $k_2$ & $k_3$ & $k_4$ & $k_5$\\
  \hline  
  \textit{Layer 1} & 0.2445 & 0.1887 & 0.1888 & 0.1889 & 0.1891\\
  \hline
  \textit{Layer 2} & 0.3421 & 0.1603 & 0.1630 & 0.1659 & 0.1688\\
  \hline
  \textit{Layer 3} & 0.2035 & 0.1615 & 0.1843 & 0.2104 & 0.2403\\
  \hline
  \textit{Layer 4} & 0.0843 & 0.1409 & 0.1877 & 0.2508 & 0.3362\\
  \hline
\end{tabular}
\caption{Kernel weights in each layer for the \textit{mnist-back-image} dataset}
\label{back_image_kw}
\end{table}
\renewcommand{\arraystretch}{1}

\section{Conclusion}
\label{conclusion}
In this paper we explored the concept of Multiple Kernel Learning in MKMs. A linear combination of multiple kernels formulated purely from unlabelled data is used in each layer of MKMs. The learning process of the proposed ML-MKL algorithm employs a greedy layerwise training for each layer. Empirical results indicates that using (unsupervised)MKL in in MKMs improves the classifier performance. The classification accuracy is better than learning machines with shallow architectures and is comparable with existing deep architectures.

The proposed ML-MKL algorithm can be extended in many ways. One immediate future direction to explore is to devise fine-tuning mechanism with respect to the entire architecture after kernel weights in each layer is obtained. Exploratory analysis on the features obtained is a second possible direction for further studies.


%

\appendices
\section{Derivation of the objective function $J(\mu)$}
\label{derivation1}
The objective function to be minimized is given by
\[ \min_{k \in K_{conv}} \frac{1}{2}\sum_{i=1}^n \norm{x_i - \sum_{x_j \in B_i} k_{ij}x_j}^2 + \gamma* \sum_{i=1}^n \sum_{x_j \in B_i} k_{ij} \norm{x_i-x_j}^2 \]
Expanding the norm on the first part of the sum
\begin{equation}
\begin{split}
\min_{k \in K_{conv}} \frac{1}{2}\sum_{i=1}^n \bigg(\norm{x_i}^2 - 2*\sum_{x_j \in B_i} k_{ij} \langle x_i, x_j \rangle + \\ k_i k_i^T \circ d_i d_i^T \circ X^TX \bigg) + \gamma* \sum_{i=1}^n \sum_{x_j \in B_i} k_{ij} \norm{x_i-x_j}^2
\end{split}
\label{new_obj}
\end{equation}
The notation `$\circ$' denotes elementwise multiplication of two vectors. Here the summation $\sum_{i=1}^n \norm{x_i}^2$ can be discarded, since it is independent of the optimization parameters. Substituting $X^TX = P$, 
\[\sum_{x_j \in B_i} k_{ij} \langle x_i, x_j \rangle = k_i \circ d_i \circ p_i\]
and 
\[ \sum_{x_j \in B_i} k_{ij} \norm{x_i-x_j}^2 = k_i \circ d_i \circ v_i \]
in \ref{new_obj} we will get the simplified objective function
\begin{equation}
\begin{split}
\min_{k \in K_{conv}} \sum_{i=1}^n \bigg( k_i k_i^T \circ d_i d_i^T \circ P + \\ 2\big(\gamma*k_i \circ v_i \circ d_i - k_i \circ p_i \circ d_i\big) \bigg)
\end{split}
\label{simplified}
\end{equation}
Here $p_i$ and $v_i$ are columns of $P$ and $M$ corresponding to $x_i$ respectively. Substituting $k_i = \sum_{t=1}^m \mu_tk_{t,i}$ in \ref{simplified} we will get
\[ \min_{\mu \in \Delta} \mu^T \Bigg( \sum_{t=1}^m \sum_{i=1}^n k_{t,i}k_{t,i}^T \circ d_i d_i^T \circ P \Bigg)^T \mu + z^T \mu \]
which is the objective function $J(\mu)$.



\ifCLASSOPTIONcaptionsoff
  \newpage
\fi

\end{document}